\documentclass[11pt,a4paper]{article}

\usepackage[margin=1in]{geometry}
\usepackage[utf8]{inputenc}
\usepackage{hyperref}
\usepackage{url}
\usepackage{natbib}
\usepackage{graphicx}
\graphicspath{{figures/}}

\begin{document}

\title{Swarm Skills: A Portable, Self-Evolving Multi-Agent System Specification for Coordination Engineering}

\author{%
  openJiuwen Team \& Gaoling School of Artificial Intelligence, Renmin University of China \\
  \texttt{research@openjiuwen.com}
}
\date{}

\maketitle

\begin{abstract}
As artificial intelligence engineering paradigms shift from single-agent Prompt and Context Engineering toward multi-agent \textbf{Coordination Engineering}, the ability to codify and systematically improve how multiple agents collaborate has emerged as a critical bottleneck. While single-agent skills can now be distributed as portable assets, multi-agent coordination protocols remain locked within framework-internal code or static configurations, preventing them from being shared across systems or autonomously improved over time. We propose \textbf{Swarm Skills}, a portable specification that extends the Anthropic Skills standard with multi-agent semantics. Swarm Skills turns multi-agent workflows into first-class, distributable assets that consist of roles, workflows, execution bounds, and a built-in semantic structure for self-evolution. To operationalize the specification's evolving nature, we present a companion self-evolution algorithm that automatically distills successful execution trajectories into new Swarm Skills and continuously patches existing ones based on multi-dimensional scoring (Effectiveness, Utilization, and Freshness), eliminating the need for human-in-the-loop oversight during the refinement process. Through an architectural compatibility analysis and a comprehensive qualitative case study using the open-source JiuwenSwarm reference implementation, we demonstrate how Swarm Skills achieves zero-adapter cross-agent portability via progressive disclosure, enabling agent teams to self-evolve their coordination strategies without framework lock-in.
\end{abstract}

\section{Introduction}

The rapid advancement of Large Language Models (LLMs) has catalyzed the evolution of AI engineering paradigms. Early efforts focused on the individual agent, moving from \textit{Prompt Engineering} to steer model outputs, to \textit{Context Engineering} for grounding responses in external knowledge, and recently to \textit{Harness Engineering} for equipping agents with robust tool-use and sandbox capabilities. However, as tasks grow in complexity—spanning domains from multi-disciplinary medical diagnosis to software engineering and architectural design—the limitations of single-agent systems have become apparent. This necessitates a shift toward \textbf{Coordination Engineering}, an emerging paradigm focused on the orchestration, task allocation, communication protocols, and isolation mechanisms required to make multiple agents collaborate effectively as a team.

A critical missing component within Coordination Engineering is \textit{coordination experience sedimentation}—the ability to capture, reuse, and iteratively improve successful multi-agent collaboration patterns. In the single-agent domain, this problem has been elegantly addressed by standards such as Anthropic Skills, which package an agent's operational knowledge into portable, distributable assets (e.g., \texttt{SKILL.md} files). Combined with single-agent self-evolution algorithms like Voyager \citep{wang2023voyager} or EvoSkills, agents can autonomously accumulate and refine their capabilities over time. 

In stark contrast, multi-agent coordination protocols currently exist either as framework-internal code (e.g., hardcoded Python scripts in AutoGen \citep{wu2023autogen}) or framework-private configuration files (e.g., CrewAI templates or MetaGPT SOPs \citep{hong2024metagpt}). When a multi-agent session concludes, the nuanced coordination decisions—such as how a complex task was decomposed, which roles were dynamically instantiated, and how conflicts were resolved—are discarded alongside the execution trajectory. The next time a similar task arises, the system must plan the collaboration from scratch. Furthermore, because these protocols are tightly coupled to their execution frameworks, they cannot be shared across different multi-agent systems, nor do they possess a native mechanism for autonomous self-improvement. We refer to the continuous improvement of collaboration patterns as \textbf{Coordination Evolution}, representing the next frontier beyond single-agent skill evolution.

To address this gap, we introduce \textbf{Swarm Skills}, a portable specification that extends the Anthropic Skills standard with multi-agent semantics. Swarm Skills elevates multi-agent coordination from ephemeral runtime state to a first-class, distributable, and self-evolving asset. By strictly separating the specification from the runtime execution environment, Swarm Skills ensures that coordination protocols can be authored once and executed by any compatible Host Agent. Furthermore, the specification natively embeds a self-evolving semantic structure (the Evolution Experience), explicitly defining how a Swarm Skill should accumulate operational feedback over time.

\begin{figure*}[t]
\centering
\includegraphics[width=0.9\textwidth]{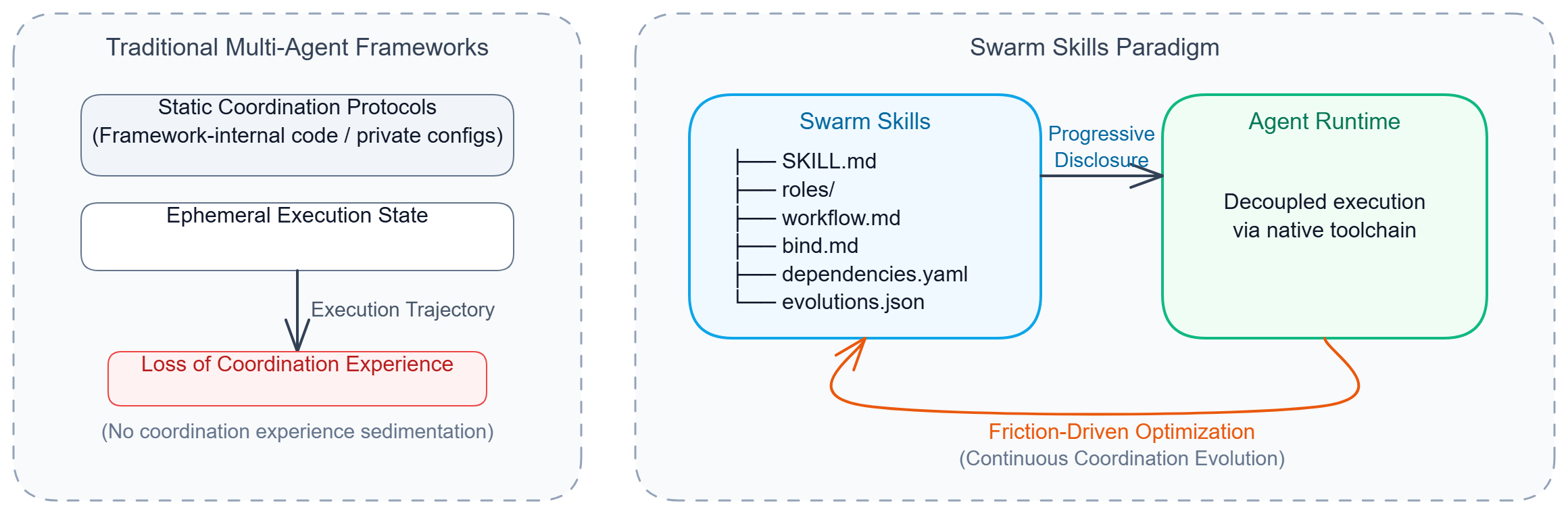}
\caption{The Paradigm Shift from Monolithic Frameworks to Portable Swarm Skills. \textbf{Left:} Traditional multi-agent frameworks embed coordination logic directly within runtime configurations, resulting in ephemeral execution traces that are discarded post-session. \textbf{Right:} The Swarm Skills paradigm abstracts coordination into a static, portable asset separated from the Agent Runtime. Through progressive disclosure and friction-driven feedback loops, the asset inherently accumulates coordination experience and self-evolves across sessions.}
\label{fig:teaser}
\end{figure*}

To operationalize the self-evolving nature of the specification, we propose a companion \textbf{Self-Evolution Algorithm}. This algorithm automatically distills successful multi-agent execution trajectories into new Swarm Skills (CREATE) and continuously optimizes existing ones (PATCH) based on friction patterns identified during execution. To prevent the accumulation of low-quality patches, the algorithm employs a multi-dimensional scoring mechanism—evaluating Effectiveness (E), Utilization (U), and Freshness (F)—coupled with automated simplification and culling routines (SIMPLIFY, REBUILD, ROLLBACK). This allows the coordination protocols to improve autonomously without requiring human-in-the-loop approval gates.

Our contributions are threefold:
\begin{enumerate}
    \item \textbf{The Swarm Skills Specification:} We propose the first multi-agent extension to the Anthropic Skills standard, comprising a five-component asset structure (frontmatter, roles, workflow, execution bounds, and dependencies) with built-in semantics for self-evolution.
    \item \textbf{A Companion Self-Evolution Algorithm:} We present an automated algorithm that operationalizes the specification's self-evolving semantics, enabling continuous, score-driven refinement of multi-agent coordination protocols without human oversight.
    \item \textbf{Architectural Compatibility \& Qualitative Validation:} Through architectural analysis, we demonstrate how Swarm Skills achieves zero-adapter cross-agent portability via progressive disclosure. We validate the specification and algorithm through a comprehensive qualitative case study (a 6-role travel planning team) using \textbf{JiuwenSwarm}\footnote{JiuwenSwarm is an AI Agent developed based on openJiuwen (\url{https://www.openjiuwen.com/}), serving as the reference implementation for Coordination Engineering and Swarm Skills. The Swarm Skills specification and community-contributed skills are accessible via the Swarm Skills Hub (\url{https://swarmskills.openjiuwen.com/}).} as the executing agent.
\end{enumerate}
\section{Related Work}

\subsection{Skill Abstractions and Single-Agent Self-Evolution}
The concept of abstracting agent capabilities into reusable skills has proven critical for scaling LLM agents. Frameworks like Voyager \citep{wang2023voyager} demonstrated that an agent can iteratively compose and refine an expanding library of executable code skills (e.g., in Minecraft) by receiving environmental feedback. Subsequently, systems like SAGE \citep{liang2024sage}, ExpeL \citep{zhao2024expel}, and EvoSkills have generalized this approach, enabling agents to self-correct and autonomously accumulate domain-specific operational knowledge. The \textbf{Anthropic Skills standard} represents a significant milestone in this trajectory, providing a standardized, distributable format (\texttt{SKILL.md}) for these single-agent capabilities. While these works successfully establish skills as portable and evolvable assets, they are fundamentally constrained to the single-agent paradigm. Our work builds directly upon the Anthropic standard, extending its portability to multi-agent environments.

\subsection{Multi-Agent Topologies and Workflow Optimization}
Significant research has explored how to structure and optimize multi-agent interactions. Foundational works such as CAMEL \citep{li2023camel}, Generative Agents \citep{park2023generative}, and ChatDev \citep{qian2024chatdev} demonstrated the potential of role-playing and communicative agents to solve complex tasks. Systems like AgentVerse \citep{chen2024agentverse}, DyLAN \citep{liu2023dylan}, and GPTSwarm \citep{zhuge2024gptswarm} frame multi-agent collaboration as a graph optimization problem, dynamically routing messages or selecting nodes to improve task performance. Similarly, AFlow \citep{zhang2025aflow} and EvoAgent \citep{yuan2025evoagent} employ automated search procedures to discover optimal workflow topologies or augment agent capabilities. At the single-agent level, frameworks integrating reasoning, acting, and tool-use---such as ReAct \citep{yao2023react}, ToolLLM \citep{qin2024toolllm}, and HuggingGPT \citep{shen2023hugginggpt}---have laid the groundwork for robust agent execution loops. However, in these systems, the resulting coordination protocol---the "who does what and when"---remains an internal data structure of the optimization algorithm. These works contribute novel \textit{search procedures} and \textit{interaction mechanisms}, whereas we contribute a \textit{portable description format} that can persist the outputs of such algorithms for cross-framework distribution.

\subsection{Existing Multi-Agent Descriptive Formats}
Current multi-agent frameworks often require users to explicitly define team structures. For example, CrewAI utilizes proprietary Python templates to define agents and sequential tasks; MetaGPT \citep{hong2024metagpt} employs Standard Operating Procedures (SOPs); and AutoGen \citep{wu2023autogen} relies on GroupChat configurations. Crucially, these descriptive formats are \textbf{framework-private}. A CrewAI template cannot be natively executed by AutoGen, nor does it possess an explicit, built-in semantic structure for self-evolution. Our specification bridges this gap by decoupling the coordination description from the execution runtime, elevating multi-agent protocols to first-class, distributable assets.

\subsection{Engineering Paradigms for LLM Agents}
The evolution of LLM applications can be viewed through successive engineering paradigms. \textit{Prompt Engineering} focuses on input optimization to elicit specific model behaviors. \textit{Context Engineering} (e.g., RAG) focuses on grounding models in external knowledge. \textit{Harness Engineering} focuses on equipping agents with robust tool-use, sandbox isolation, and sequential execution loops. We position our work within \textbf{Coordination Engineering}, an emerging paradigm proposed by the openJiuwen community that focuses on orchestrating multi-agent systems—encompassing team formation, communication protocols, and fault recovery. Within this paradigm, Swarm Skills specifically addresses the critical challenge of \textit{coordination experience sedimentation}.

\section{Motivation: Multi-Agent Skill Demand in the Wild}
To validate the necessity of a standardized multi-agent skill format, we conducted a measurement study of existing community repositories utilizing the Anthropic Skills standard (crawled in April 2026). Our analysis of 33 search queries across 9 skill repositories revealed a strong "revealed preference" for multi-agent capabilities.

First, power users frequently attempt to simulate multi-agent dynamics using single-agent \texttt{SKILL.md} files. We identified numerous highly downloaded skills explicitly named using team-style nomenclature (e.g., \texttt{c-level-advisor} containing 9 distinct roles, \texttt{ra-qm-team} defining 13 expert personas, and \texttt{engineering-team} specifying 14 roles). Second, skills designed for adversarial or parallel workflows exhibit disproportionately high adoption rates (e.g., \texttt{critique} skills with $>$77.8K installs, \texttt{parallel-debugging}, \texttt{adversarial-reviewer}). 

These findings indicate that users are actively employing workarounds to express multi-role collaboration requirements within a single-agent format. This strongly supports the premise that multi-agent coordination protocols—and the ability to share them—represent a pressing demand within the Coordination Engineering paradigm.

\section{Discussion, Limitations, and Future Work}
In this specification paper, we introduced Swarm Skills, a portable format for defining and evolving multi-agent coordination protocols, supported by a self-evolution algorithm. While the architectural analysis and case study demonstrate the feasibility of zero-adapter deployment and autonomous refinement via JiuwenSwarm, several challenges remain.

\textbf{Limitations:} The primary limitation of this work is the absence of large-scale, quantitative conformance testing across diverse Host Agents (e.g., AutoGen, LangGraph). Because Swarm Skills relies on progressive disclosure, its effectiveness on Host Agents that lack native recursive \texttt{read\_file} or dynamic sub-agent spawning capabilities remains to be empirically evaluated. Furthermore, evaluating the quality of multi-agent coordination lacks standardized benchmarks, making it difficult to rigorously quantify the performance gains provided by the self-evolution algorithm across different domains.

\textbf{Future Work:} Our immediate future work will focus on extensive empirical conformance tests across commercial and open-source multi-agent frameworks. Additionally, we aim to explore more sophisticated governance mechanisms for resolving "first-run lock-in"—scenarios where a severely suboptimal initial workflow becomes heavily patched rather than fundamentally restructured. Finally, while Swarm Skills addresses the \textit{sedimentation} of coordination experience, the broader \textbf{Coordination Engineering} paradigm contains numerous open challenges regarding fault tolerance, optimal message routing, and inter-agent isolation that require continued community collaboration.
\section{The Swarm Skills Specification}

At its core, Swarm Skills is a specification designed around a single philosophy: \textbf{a description format, not a runtime}. The specification dictates the file structure and semantics of a multi-agent coordination protocol, entirely decoupled from the framework-specific implementation of message routing, state management, or sub-agent spawning. Any compliant \textbf{Host Agent} can load a Swarm Skill and execute the described workflow using its native toolchain. By strictly separating the declaration of coordination logic from its execution, Swarm Skills establishes a universally portable asset format. Furthermore, the specification inherently supports continuous refinement by codifying self-evolving semantics directly into the asset's structure.

\subsection{Specification Architecture and the "Five Components"}
A Swarm Skill is distributed as a directory containing five required author-provided components, alongside an automatically generated runtime artifact that captures the skill's evolution experience. This modular structure adheres to the principle of \textit{Progressive Disclosure}, a design pattern championed by the Anthropic Skills standard, wherein the Host Agent initially loads only the minimum required metadata and subsequently fetches detailed instructions as needed.

The structure of a Swarm Skill is as follows:
\begin{figure*}[t]
\centering
\includegraphics[width=\textwidth]{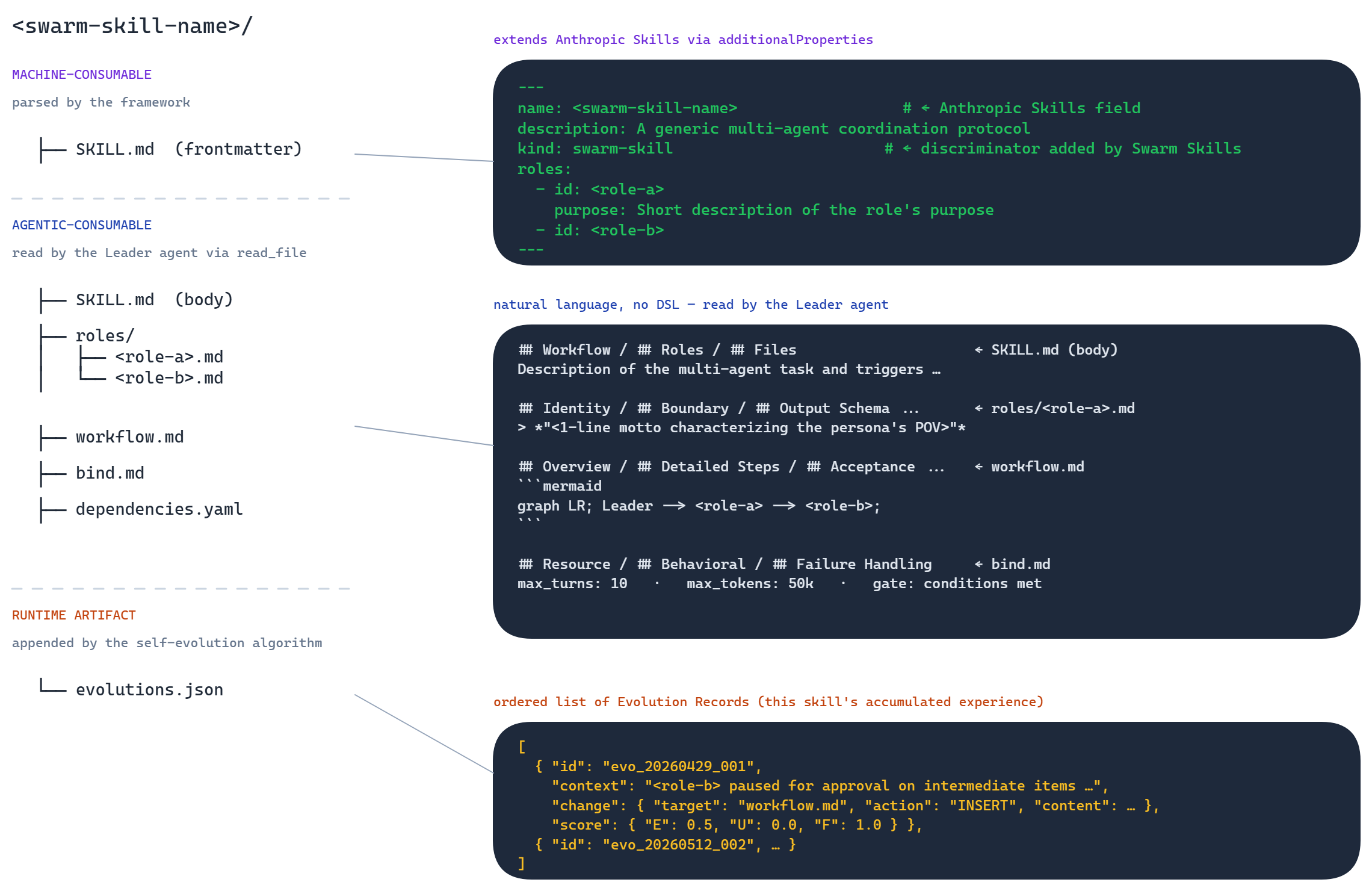}
\caption{The Anatomy of a Swarm Skill. The specification delineates the asset into three distinct functional categories: machine-consumable metadata for environment orchestration, agentic-consumable natural language instructions for the executing agents, and a dynamically appended runtime artifact (\texttt{evolutions.json}) that captures the skill's Evolution Experience.}
\label{fig:anatomy}
\end{figure*}

\subsubsection{Frontmatter Design for Mechanical Consumption}
The entry point of every Swarm Skill is \texttt{SKILL.md}. The file begins with a YAML frontmatter block designed for \textit{mechanical consumption} by the Host Agent's framework code. This block extends the standard Anthropic Skills frontmatter by introducing multi-agent semantics:

\begin{itemize}
    \item \texttt{kind}: Set to \texttt{swarm-skill}. This discriminator enables the framework to differentiate between a standard single-agent skill and a Swarm Skill, triggering the appropriate multi-agent execution pathways (e.g., configuring a Leader agent context).
    \item \texttt{teammate\_mode}: Specifies the overarching interaction paradigm for the team (e.g., \texttt{build\_mode} for autonomous execution versus \texttt{plan\_mode} for workflows requiring explicit plan approval before action).
    \item \texttt{roles[]}: An array defining the environment configuration for each participant. Each entry specifies a unique \texttt{id}, the specific \texttt{skills} and \texttt{tools} required by that role, and optionally a target \texttt{model}.
\end{itemize}

This design choice ensures that the framework code only parses the metadata necessary for orchestrating the environment. Because the Anthropic specification permits \texttt{additionalProperties: true}, Host Agents that do not support multi-agent execution will gracefully ignore the \texttt{kind} and \texttt{roles[]} fields, treating the Swarm Skill as a standard single-agent asset.

\subsubsection{Natural Language Body for Agentic Consumption}
Beneath the frontmatter, the body of \texttt{SKILL.md} is authored in natural language, intended for \textit{agentic consumption}. It provides the Leader agent with a high-level overview of the workflow, the trigger conditions for the skill, and explicit references to the supplementary files (\texttt{roles/}, \texttt{workflow.md}, and \texttt{bind.md}). 

By leveraging the Host Agent's native \texttt{read\_file} capability, the Leader autonomously retrieves the detailed personas from the \texttt{roles/} directory when instantiating Teammate agents. The \texttt{workflow.md} file describes the task dependency graph (e.g., sequential handoffs, parallel execution, or fan-out/fan-in patterns) using natural language or Mermaid syntax. Finally, \texttt{bind.md} establishes the operational boundaries, such as maximum message turns, token budgets, and explicit quality gates that must be satisfied before task completion.

\subsection{Embedded Self-Evolving Semantics}
A critical innovation of the Swarm Skills specification is the explicit inclusion of self-evolving semantics as a contractual obligation of the format. Multi-agent workflows are highly susceptible to friction—such as redundant communication, ambiguous role boundaries, or suboptimal task sequencing. A static specification cannot adapt to these runtime inefficiencies.

To address this, Swarm Skills mandates the presence of an \texttt{evolutions.json} file, which serves as the skill's \textbf{Evolution Experience}. This file stores an array of \textbf{Evolution Records}, representing discrete patches to the skill's workflow or role definitions.

The specification defines the schema for an Evolution Record as follows:
\begin{itemize}
    \item \textbf{Context:} A description of the friction encountered during a specific execution trajectory (e.g., "The Reviewer and Architect repeatedly conflicted over the sequencing of their analysis").
    \item \textbf{Change Directive:} The targeted modification to the asset (e.g., "INSERT into \texttt{workflow.md}: Add a preliminary alignment phase before parallel review").
    \item \textbf{Scoring Metrics:} Embedded fields for Effectiveness ($E$), Utilization ($U$), and Freshness ($F$), allowing the Host Agent to evaluate the historical utility of the patch.
\end{itemize}

By codifying the Evolution Experience into the specification, Swarm Skills guarantees that coordination improvements are inherently portable. If a Swarm Skill is transferred from Host Agent A to Host Agent B, the accumulated Evolution Records are transferred alongside the base protocol, ensuring that Agent B benefits from the coordination lessons learned by Agent A.

\subsection{Design Trade-offs and the Boundary of Specification}
The primary design tension in standardizing multi-agent collaboration lies in determining what to specify and what to delegate. The Swarm Skills specification intentionally omits the mechanics of message passing, task queues, and process isolation. These are considered implementation details of the Host Agent's runtime environment.

For example, whether the Host Agent implements communication via a shared whiteboard, asynchronous message queues, or direct prompt injection is irrelevant to the Swarm Skill. The specification simply declares \textit{who} needs to collaborate, \textit{what} dependencies exist between their tasks, and the \textit{constraints} governing their execution. This deliberate restraint maximizes portability across fundamentally different architectures—ranging from true multi-process multi-agent systems to single-agent systems utilizing a sub-agent (\texttt{Task}) tool. By extending an existing, widely adopted standard (Anthropic Skills) rather than inventing a bespoke domain-specific language (DSL), Swarm Skills minimizes the barrier to entry and ensures zero-adapter compatibility with the current ecosystem.
\section{A Self-Evolution Algorithm for Swarm Skills}

The Swarm Skills specification establishes the structural prerequisites for multi-agent capability distribution, notably introducing an explicit `evolutions.json` artifact. To demonstrate how this format enables autonomous refinement over time, we present a \textbf{Self-Evolution Algorithm} designed to operationalize these self-evolving semantics. This algorithm provides a reference mechanism for a Host Agent to extract coordination protocols from raw execution trajectories, patch existing workflows based on runtime friction, and curate accumulated improvements using an automated, multi-dimensional scoring system.

Importantly, while this algorithm realizes the potential of the specification, \textbf{it is not strictly required by the standard}. The specification remains agnostic; any Host Agent is free to implement alternative evolution mechanisms, provided they conform to the underlying Evolution Experience schema.

\subsection{The Three-Stage Lifecycle}
The algorithm manages the lifecycle of a Swarm Skill across three distinct stages: \textbf{CREATE}, \textbf{USE}, and \textbf{PATCH}, forming a continuous loop of capability acquisition and refinement.

\begin{figure*}[t]
\centering
\includegraphics[width=\textwidth]{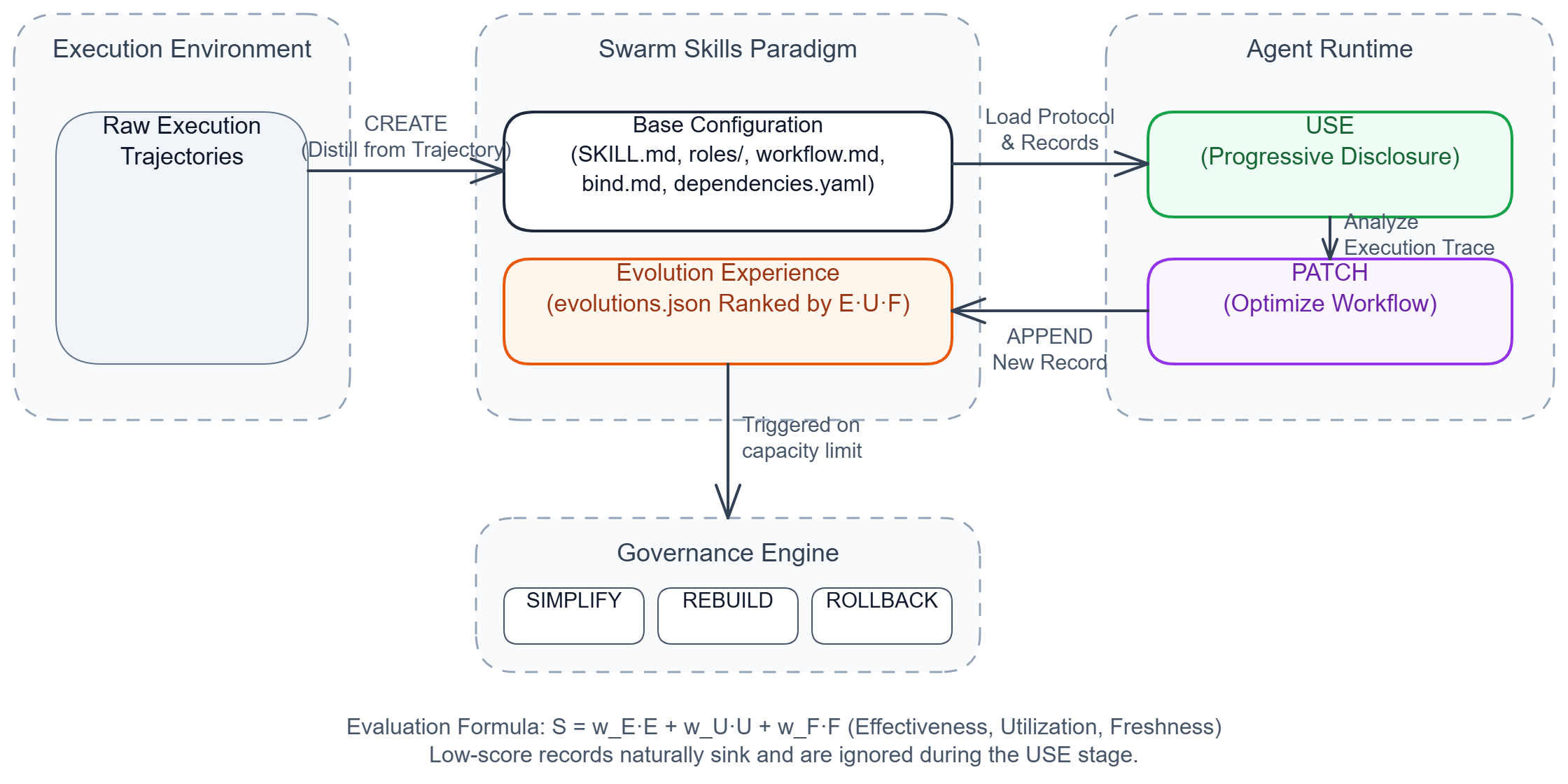}
\caption{The Self-Evolution Lifecycle. The algorithm orchestrates a continuous loop starting from Trajectory Distillation (\textbf{CREATE}). During execution (\textbf{USE}), progressive disclosure securely loads protocols and historical records. Post-execution Friction Analysis (\textbf{PATCH}) appends new records. A Governance Engine periodically simplifies, rebuilds, or rolls back the asset based on the multidimensional $S=w_E \cdot E + w_U \cdot U + w_F \cdot F$ scoring of accumulated records.}
\label{fig:lifecycle}
\end{figure*}

\subsubsection{CREATE: Trajectory Distillation}
Multi-agent collaboration patterns frequently emerge organically before they are formally codified. A Host Agent executing a complex task may dynamically instantiate multiple sub-agents, defining their personas and task dependencies on the fly. 

The \textbf{CREATE} stage monitors these interactions. A \textit{trajectory aggregator} identifies multi-agent collaboration signals (e.g., the instantiation of $\geq 2$ distinct sub-agent roles and the presence of cross-agent task dependencies). Upon detecting a sufficiently complex execution, an LLM-based distillation process synthesizes the raw trajectory into a candidate Swarm Skill. This involves:
\begin{enumerate}
    \item Abstracting the task objective into a generalized workflow pattern (e.g., transforming "Analyze Framework A vs. B" into "N-way Component Comparison").
    \item Extracting the observed persona configurations and constraints into discrete \texttt{roles/} and \texttt{bind.md} files.
    \item Storing the candidate Swarm Skill in a staging directory. Its ultimate adoption is governed entirely by the Host Agent's subsequent selection behavior during the \textbf{USE} stage.
\end{enumerate}

\subsubsection{USE: Progressive Disclosure and Agentic Routing}
When a new task arrives, the Host Agent retrieves the \texttt{description} field from the frontmatter of all available Swarm Skills and injects them into the Leader agent's system prompt. This exposes the available coordination capabilities without consuming the token budget required for their full implementations.

If the Leader selects a Swarm Skill, it utilizes the Host Agent's native \texttt{read\_file} capability to incrementally load the \texttt{SKILL.md} body, followed by the required \texttt{roles/}, \texttt{workflow.md}, and \texttt{bind.md} definitions. The Leader then orchestrates the specified team structure. Throughout this process, any applicable Evolution Records from the \texttt{evolutions.json} file are appended to the relevant instructions, ensuring that the team benefits from historical optimizations.

\subsubsection{PATCH: Friction-Driven Optimization}
The core engine of self-evolution is the \textbf{PATCH} mechanism. Upon the completion of a multi-agent task utilizing a Swarm Skill, a \textit{trajectory analyzer} scans the execution trace for two distinct signals:
\begin{itemize}
    \item \textbf{Implicit Friction Patterns:} Instances where the coordination protocol degraded (e.g., circular dependencies, redundant communication loops, or premature task termination).
    \item \textbf{Explicit Improvement Signals:} Direct user feedback or internal agent reflections suggesting workflow modifications.
\end{itemize}

If optimization opportunities are identified, the analyzer generates a new \textbf{Evolution Record} and automatically appends it to the Swarm Skill's \texttt{evolutions.json} file. This ensures that the base specification files (\texttt{SKILL.md}, \texttt{roles/}, \texttt{workflow.md}) remain untouched, mitigating the risk of catastrophic degradation from a single erroneous patch.

\subsection{Multi-Dimensional Scoring and Automated Culling}
As a Swarm Skill undergoes repeated execution, the accumulation of Evolution Records can result in prompt bloat and conflicting instructions. To enforce rigorous quality control without human intervention, the algorithm implements a multi-dimensional scoring system that continuously evaluates each Evolution Record.

The total score $S_i$ for an Evolution Record $i$ is defined as a weighted composite of three metrics: \textbf{Effectiveness ($E$)}, \textbf{Utilization ($U$)}, and \textbf{Freshness ($F$)}.

\begin{enumerate}
    \item \textbf{Effectiveness ($E \in [0, 1]$):} Measures the qualitative impact of the patch. The algorithm employs Bayesian smoothing with a Beta(1,1) prior to stabilize early-stage evaluations, updating the score based on the outcome of tasks executing the patch.
    \item \textbf{Utilization ($U \in [0, 1]$):} The adoption rate of the patch. If the Host Agent's Leader consistently ignores an appended instruction, $U$ decays, indicating low utility.
    \item \textbf{Freshness ($F \in [0, 1]$):} A time-decay factor applying an exponential half-life (e.g., 90 days), ensuring that stale optimizations are gradually deprecated in favor of recent adaptations.
\end{enumerate}

Records whose composite score $S_i$ falls below a predetermined threshold naturally sink to the bottom of the context window and are eventually ignored by the Host Agent. 

\subsection{Governance Actions: SIMPLIFY, REBUILD, and ROLLBACK}
To maintain the long-term structural integrity of the Swarm Skill, the algorithm provides three governance routines, triggered automatically when the \texttt{evolutions.json} array exceeds a specified capacity (e.g., $\geq 10$ records):

\begin{itemize}
    \item \textbf{SIMPLIFY:} Executes an LLM-driven pruning operation on the Evolution Experience, categorizing records for deletion (low score/stale), merging (overlapping intent), refinement (clarity), or retention.
    \item \textbf{REBUILD:} Addresses "first-run lock-in," where a suboptimal initial workflow becomes heavily patched. This action fundamentally rewrites the core \texttt{SKILL.md} and supplementary files, incorporating the accumulated patches into the base specification. The previous version is archived, and the \texttt{evolutions.json} array is cleared.
    \item \textbf{ROLLBACK:} Provides a fail-safe mechanism to revert a Swarm Skill to any previously archived state following a detrimental REBUILD operation.
\end{itemize}

\subsection{The JiuwenSwarm Reference Implementation}
The self-evolution algorithm, including the CREATE, PATCH, and governance routines, has been implemented within \textbf{JiuwenSwarm}, an open-source reference agent developed by the openJiuwen community. Because the algorithm strictly interacts with the schema defined by the Swarm Skills specification, it requires no internal modifications to the underlying multi-agent execution framework. This demonstrates that the operationalization of self-evolving semantics is itself a portable capability.

\section{Architectural Compatibility \& Deployment}
A critical objective of the Swarm Skills specification is achieving \textbf{Zero-Adapter Deployment}. In this section, we provide a theoretical and architectural analysis demonstrating how Swarm Skills can be deployed across heterogeneous multi-agent frameworks without requiring translation layers or framework-specific plugins.

\subsection{Compatibility via Progressive Disclosure}
The architectural portability of Swarm Skills is rooted in its strict adherence to the Anthropic Skills standard, specifically the principle of \textbf{Progressive Disclosure}. 

Traditional multi-agent frameworks (e.g., CrewAI, MetaGPT) define their topologies using bespoke, internal Data Structures (e.g., Python classes or proprietary YAML configurations). Transferring a CrewAI template to AutoGen requires a translation adapter to map the configuration schemas.

Conversely, Swarm Skills relies exclusively on natural language and universally accepted formats (Markdown). The only structured metadata resides in the \texttt{SKILL.md} frontmatter. When deployed to an Anthropic-compatible \textbf{Host Agent} (e.g., Claude Code), the integration process is as simple as copying the Swarm Skill directory into the Host's \texttt{skills/} folder. The Host Agent's native \texttt{read\_file} tool handles the recursive loading of the \texttt{roles/} and \texttt{workflow.md} files, while its native sub-agent spawning tool (e.g., \texttt{Task}) instantiates the required Teammates. The entire multi-agent orchestration is driven by the Host Agent's ability to comprehend the natural language workflow, not by its ability to parse a proprietary execution graph.

\subsection{Graceful Degradation}
The specification mandates that multi-agent configuration variables (e.g., the \texttt{roles[]} array) are located within the frontmatter. Because the underlying Anthropic specification permits \texttt{additionalProperties: true}, Host Agents that lack multi-agent capabilities (e.g., standard single-agent LLM wrappers) will not fail upon encountering a Swarm Skill. 

Instead, these systems gracefully degrade: they ignore the \texttt{kind: swarm-skill} discriminator and the \texttt{roles[]} array, treating the directory as a standard, single-agent Skill. The single agent will attempt to execute the workflow sequentially, adopting the various personas as needed. This guarantees backward compatibility across the entire spectrum of LLM execution environments, satisfying the requirement for a truly portable asset.

While comprehensive, large-scale empirical conformance testing across multiple commercial frameworks remains a focus for future work, the architectural alignment with established progressive disclosure mechanics strongly supports the universal portability of the Swarm Skills specification.
\section{Qualitative Case Study: The \texttt{travel-planning-swarm} Skill}

To provide a concrete demonstration of the Swarm Skills specification in practice and illustrate the mechanics of the self-evolution algorithm, we conduct a qualitative system walkthrough using a \texttt{travel-planning-swarm} Swarm Skill. This case study utilizes the JiuwenSwarm reference implementation to orchestrate a complex, multi-constraint planning task.

\subsection{Design Rationale and Initial Setup}
Travel planning for a family involves highly coupled constraints---logistics, accommodations, activities, and budget---that often conflict. To model this, the initial \texttt{travel-planning-swarm} Skill was structured with five distinct expert roles:
\begin{itemize}
    \item \textbf{Transportation Expert:} Responsible for flights, trains, rental cars, and local transit, focusing on price comparison and scheduling.
    \item \textbf{Accommodation Expert:} Focused on booking hotels suitable for specific demographics and locations.
    \item \textbf{Attraction Expert:} Responsible for designing daily itineraries and ticketing.
    \item \textbf{Plan Synthesizer:} Tasked with aggregating individual expert plans into a cohesive, high cost-performance proposal.
    \item \textbf{Budget Reviewer:} Responsible for ensuring the total cost remains within the specified limit and drafting a final sharing post for social media (e.g., WeChat Moments).
\end{itemize}

The user provides a complex prompt: \textit{"Plan a May Day holiday trip for my wife, my 1-year-old child, and me, traveling from Hangzhou to Northeast China with a total budget of 20,000 RMB."} Upon receiving this request, the Host Agent utilizes the \texttt{travel-planning-swarm} skill to instantiate the dedicated expert team.

\subsection{System Walkthrough: Autonomous Conflict Resolution}

\begin{figure*}[t]
\centering
\includegraphics[width=0.9\textwidth]{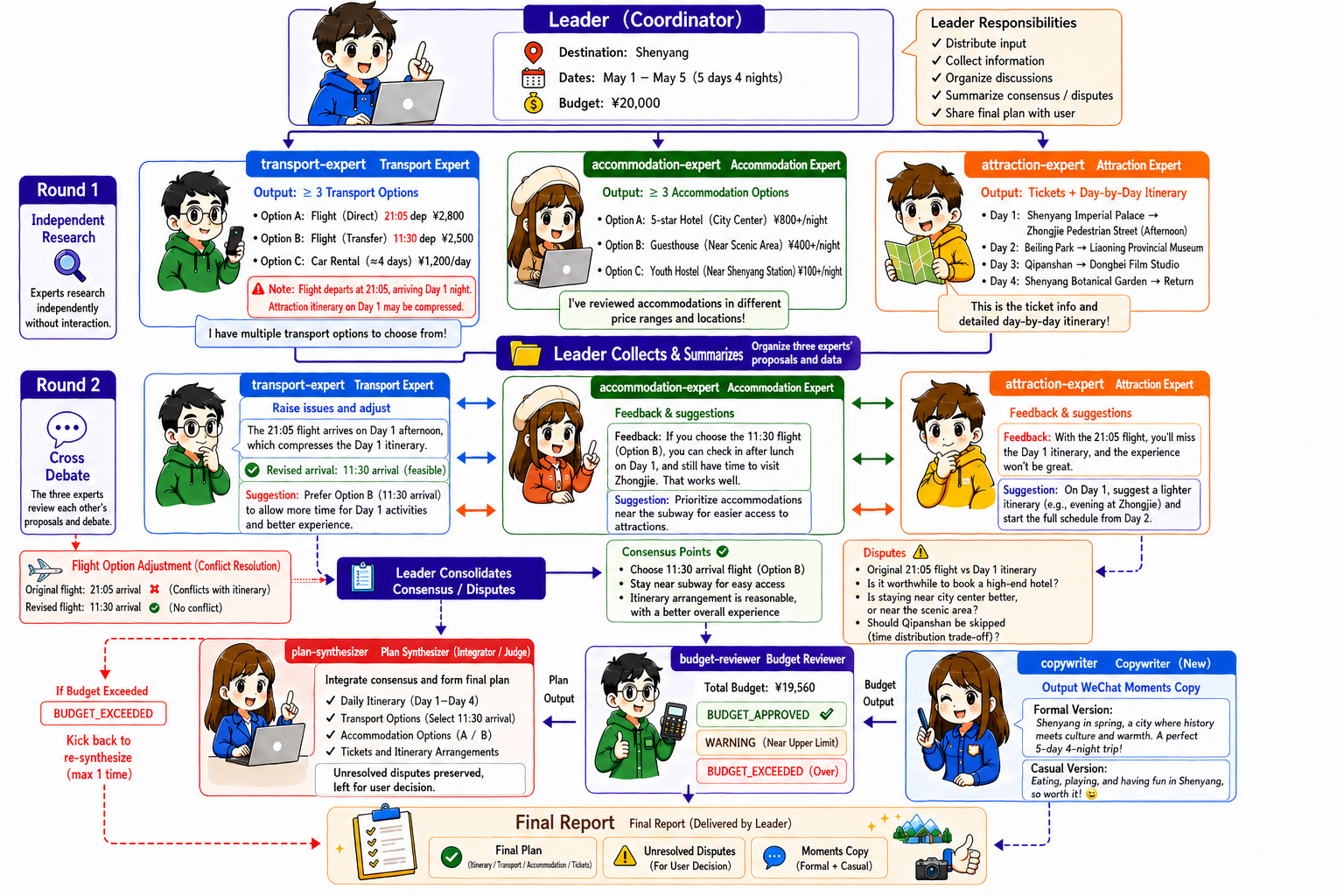}
\caption{The overall collaboration workflow of the \texttt{travel-planning-swarm} Swarm Skill. The diagram illustrates the dynamic execution phase, including parallel task processing, autonomous conflict resolution between domain experts, and the final synthesis and review stages.}
\label{fig:workflow}
\end{figure*}

During the execution phase (as illustrated in Figure \ref{fig:workflow}), the experts operate concurrently to draft their respective domain plans, adhering to the constraints defined in their \texttt{roles/<id>.md} files. The \texttt{workflow.md} mandates that experts must cross-reference their proposals and resolve temporal or spatial conflicts before submitting them to the Plan Synthesizer.

A conflict arises during this phase: The Transportation Expert initially selects a cost-effective evening flight arriving in the Northeast. Simultaneously, the Attraction Expert, prioritizing maximizing vacation time, schedules an afternoon sightseeing itinerary for the first day. Recognizing the temporal impossibility of this overlap, the two agents autonomously initiate a negotiation dialogue. Weighing the constraints of traveling with a 1-year-old child against the budget, they collaboratively decide to update the flight to a morning departure, ensuring the afternoon itinerary remains viable while keeping the overall cost strictly within the 20,000 RMB limit. The Plan Synthesizer then successfully aggregates the reconciled sub-plans into a final itinerary.

\subsection{Demonstrating the Evolution Experience}
While the team successfully delivered the travel plan, the post-execution analysis by the self-evolution algorithm identified a structural inefficiency. The system detected a clear responsibility coupling (role friction) within the \textbf{Budget Reviewer} role, which was tasked with both rigorous financial auditing and creative social media copywriting.

The self-evolution algorithm automatically generated an Evolution Record suggesting a structural split to decouple these responsibilities, and submitted it for user approval\footnote{While the core algorithm described in Section 5 can operate fully autonomously, the JiuwenSwarm deployment exposes an optional human-review step (interactive mode) for structural changes to ensure operational safety and user alignment.}.

\begin{verbatim}
{
  "id": "evo_20260430_001",
  "context": "The Budget Reviewer experienced context switching and latency 
              when transitioning from numerical auditing to creative writing.",
  "change_directive": {
    "target_files": ["roles/copywriter.md", "workflow.md"],
    "action": "SPLIT_ROLE",
    "content": "Create a dedicated 'Copywriting Expert' for social media 
                sharing. Update workflow to execute copywriting in parallel 
                with final budget review."
  },
  "metrics": {
    "effectiveness_score": 0.5,
    "utilization_rate": 0.0,
    "freshness_decay": 1.0
  }
}
\end{verbatim}

Upon user confirmation of the evolution experience, the system executed a "one-click rebuild" (REBUILD phase). It automatically generated a new \texttt{roles/copywriter.md} file, instantiating the sixth member, the \textbf{Copywriting Expert}, and patched the \texttt{workflow.md} to route the final approved plan to this new role for parallel processing. 

Simultaneously, the nuanced negotiation tactics employed by the Transportation and Attraction experts (e.g., prioritizing infant travel comfort over minor cost savings) were distilled and appended as behavioral patches to their respective role files.

This case study illustrates the core value proposition of Swarm Skills: the team structure automatically adjusts, the collaboration workflow is continuously optimized, and individual member experience is progressively accumulated. Ultimately, the more the Swarm Skill is utilized, the stronger and more efficient the agent team becomes.

\bibliographystyle{plain}
\bibliography{references}

@article{wang2023voyager,
  author       = {Guanzhi Wang and
                  Yuqi Xie and
                  Yunfan Jiang and
                  Ajay Mandlekar and
                  Chaowei Xiao and
                  Yuke Zhu and
                  Linxi Fan and
                  Anima Anandkumar},
  title        = {Voyager: An Open-Ended Embodied Agent with Large Language Models},
  journal      = {Trans. Mach. Learn. Res.},
  volume       = {2024},
  year         = {2024},
  url          = {https://openreview.net/forum?id=ehfRiF0R3a},
  bibsource    = {dblp computer science bibliography, https://dblp.org}
}

@inproceedings{wu2023autogen,
  title={AutoGen: Enabling Next-Gen LLM Applications via Multi-Agent Conversation},
  author={Wu, Qingyun and Bansal, Gagan and Zhang, Jieyu and Wu, Yiran and Li, Beibin and Zhu, Erkang and Jiang, Li and Zhang, Xiaoyun and Zhang, Shaokun and Liu, Jiale and Awadallah, Ahmed Hassan and White, Ryen W and Burger, Doug and Wang, Chi},
  booktitle={First Conference on Language Modeling},
  year={2024}
}

@inproceedings{hong2024metagpt,
  title={{MetaGPT}: Meta Programming for A Multi-Agent Collaborative Framework},
  author={Hong, Sirui and Zhuge, Mingchen and Chen, Jonathan and Zheng, Xiawu and Cheng, Yuheng and Wang, Jinlin and Zhang, Ceyao and Wang, Zili and Yau, Steven Ka Shing and Lin, Zijuan and Zhou, Liyang and Ran, Chenyu and Xiao, Lingfeng and Wu, Chenglin and Schmidhuber, J{\"u}rgen},
  booktitle={The Twelfth International Conference on Learning Representations},
  year={2024}
}

@article{liang2024sage,
  title={{SAGE}: Self-evolving Agents with Reflective and Memory-augmented Abilities},
  author={Liang, Xuechen and He, Yangfan and Xia, Yinghui and Song, Xinyuan and Wang, Jianhui and Tao, Meiling and Sun, Li and Yuan, Xinhang and Su, Jiayi and Li, Keqin and Chen, Jiaqi and Yang, Jinsong and Chen, Siyuan and Shi, Tianyu},
  journal={Neurocomputing},
  year={2025}
}

@inproceedings{zhao2024expel,
  title={{ExpeL}: {LLM} Agents Are Experiential Learners},
  author={Zhao, Andrew and Huang, Daniel and Xu, Quentin and Lin, Matthieu and Liu, Yong-Jin and Huang, Gao},
  booktitle={Proceedings of the AAAI Conference on Artificial Intelligence},
  year={2024}
}

@inproceedings{liu2023dylan,
  title={A Dynamic {LLM}-Powered Agent Network for Task-Oriented Agent Collaboration},
  author={Liu, Zijun and Zhang, Yanzhe and Li, Peng and Liu, Yang and Yang, Diyi},
  booktitle={First Conference on Language Modeling},
  year={2024}
}

@inproceedings{zhuge2024gptswarm,
  title={{GPTSwarm}: Language Agents as Optimizable Graphs},
  author={Zhuge, Mingchen and Wang, Wenyi and Kirsch, Louis and Faccio, Francesco and Khizbullin, Dmitrii and Schmidhuber, J{\"u}rgen},
  booktitle={Forty-first International Conference on Machine Learning},
  year={2024}
}

@inproceedings{zhang2025aflow,
  title={{AFlow}: Automating Agentic Workflow Generation},
  author={Zhang, Jiayi and Xiang, Jinyu and Yu, Zhaoyang and Teng, Fengwei and Chen, Xionghui and Chen, Jiaqi and Zhuge, Mingchen and Cheng, Xin and Hong, Sirui and Wang, Jinlin and Zheng, Bingnan and Liu, Bang and Luo, Yuyu and Wu, Chenglin},
  booktitle={The Thirteenth International Conference on Learning Representations},
  year={2025}
}

@inproceedings{yuan2025evoagent,
  title={{EvoAgent}: Towards Automatic Multi-Agent Generation via Evolutionary Algorithms},
  author={Yuan, Siyu and Song, Kaitao and Chen, Jiangjie and Tan, Xu and Li, Dongsheng and Yang, Deqing},
  booktitle={Proceedings of the 2025 Conference of the Nations of the Americas Chapter of the Association for Computational Linguistics (NAACL)},
  year={2025}
}

@inproceedings{li2023camel,
  title={{CAMEL}: Communicative Agents for ``Mind'' Exploration of Large Language Model Society},
  author={Li, Guohao and Hammoud, Hasan Abed Al Kader and Itani, Hani and Khizbullin, Dmitrii and Ghanem, Bernard},
  booktitle={Thirty-seventh Conference on Neural Information Processing Systems},
  year={2023}
}

@inproceedings{qian2024chatdev,
  title={{ChatDev}: Communicative Agents for Software Development},
  author={Qian, Chen and Liu, Wei and Liu, Hongzhang and Chen, Nuo and Dang, Yufan and Li, Jiahao and Yang, Cheng and Chen, Weize and Su, Yusheng and Cong, Xin and Xu, Juyuan and Li, Dahai and Liu, Zhiyuan and Sun, Maosong},
  booktitle={Proceedings of the 62nd Annual Meeting of the Association for Computational Linguistics (ACL)},
  year={2024}
}

@inproceedings{park2023generative,
  title={Generative Agents: Interactive Simulacra of Human Behavior},
  author={Park, Joon Sung and O'Brien, Joseph C. and Cai, Carrie J. and Morris, Meredith Ringel and Liang, Percy and Bernstein, Michael S.},
  booktitle={Proceedings of the 36th Annual ACM Symposium on User Interface Software and Technology (UIST)},
  pages={1--22},
  year={2023}
}

@inproceedings{yao2023react,
  title={{ReAct}: Synergizing Reasoning and Acting in Language Models},
  author={Yao, Shunyu and Zhao, Jeffrey and Yu, Dian and Du, Nan and Shafran, Izhak and Narasimhan, Karthik R. and Cao, Yuan},
  booktitle={The Eleventh International Conference on Learning Representations},
  year={2023}
}

@inproceedings{chen2024agentverse,
  title={{AgentVerse}: Facilitating Multi-Agent Collaboration and Exploring Emergent Behaviors},
  author={Chen, Weize and Su, Yusheng and Zuo, Jingwei and Yang, Cheng and Yuan, Chenfei and Chan, Chi-Min and Yu, Heyang and Lu, Yaxi and Hung, Yi-Hsin and Qian, Chen and Qin, Yujia and Cong, Xin and Xie, Ruobing and Liu, Zhiyuan and Sun, Maosong and Zhou, Jie},
  booktitle={The Twelfth International Conference on Learning Representations},
  year={2024}
}

@inproceedings{qin2024toolllm,
  title={{ToolLLM}: Facilitating Large Language Models to Master 16000+ Real-world APIs},
  author={Qin, Yujia and Liang, Shihao and Ye, Yining and Zhu, Kunlun and Yan, Lan and Lu, Yaxi and Lin, Yankai and Cong, Xin and Tang, Xiangru and Qian, Bill and Zhao, Sihan and Hong, Lauren and Tian, Runchu and Xie, Ruobing and Zhou, Jie and Gerstein, Mark and Li, Dahai and Liu, Zhiyuan and Sun, Maosong},
  booktitle={The Twelfth International Conference on Learning Representations},
  year={2024}
}

@inproceedings{shen2023hugginggpt,
  title={{HuggingGPT}: Solving {AI} Tasks with {ChatGPT} and its Friends in {Hugging Face}},
  author={Shen, Yongliang and Song, Kaitao and Tan, Xu and Li, Dongsheng and Lu, Weiming and Zhuang, Yueting},
  booktitle={Advances in Neural Information Processing Systems},
  volume={36},
  year={2023}
}

\appendix
\section{Author List}
\label{app:authors}

\paragraph{Core Contributors.}
Xinyu Zhang, Zhicheng Dou, Deyang Li, Jianjun Tao, Shuo Cheng, Ruifeng Shi, Fangchao Liu, Enrui Hu, Yangkai Ding, Hongbo Wang, Qi Ye, Xuefeng Jin, Zhangchun Zhao

\end{document}